\documentclass[letterpaper, 10 pt, conference]{ieeeconf-modified}
\IEEEoverridecommandlockouts
\usepackage{graphicx}
\graphicspath{ {./figures/} }
\usepackage{amsmath}

\usepackage{mathtools}

\usepackage[numbers]{natbib}

\usepackage{textcomp}
\usepackage{array}
\usepackage{multirow}
\usepackage[bottom]{footmisc}
\usepackage{hyperref}
\usepackage[capitalize]{cleveref}

\crefname{equation}{}{}

\title{\LARGE \bf
Learning Purely Tactile In-Hand Manipulation with a Torque-Controlled Hand}
\author{Leon Sievers*, Johannes Pitz* and Berthold Bäuml
	\thanks{
		\hspace{-5mm}
		First two authors contributed equally. All authors are affiliated with the Institute of Robotics and Mechatronics, German Aerospace Center (DLR) and the Deggendorf Institute of Technology, Germany. }
		\thanks{\hspace{-5mm}\scriptsize\texttt{\{leon.sievers, johannes.pitz, berthold.baeuml\}@dlr.de}}
	\thanks{
		\hspace{-5mm}
		Website: \scriptsize\url{dlr-alr.github.io/dlr-tactile-manipulation}
	}
}

\begin{document}

\maketitle
\thispagestyle{empty}
\pagestyle{empty}

\begin{abstract}
	We show that a purely tactile dextrous in-hand manipulation task with continuous regrasping, requiring permanent force closure, can be learned from scratch and executed robustly on a torque-controlled humanoid robotic hand.
	The task is rotating a cube without dropping it, but in contrast to OpenAI's seminal cube manipulation task~\cite{OpenAI2018}, the palm faces downwards and no cameras but only the hand's position and torque sensing are used.
	Although the task seems simple, it combines for the first time all the challenges in execution as well as learning that are important for using in-hand manipulation in real-world applications.
	We efficiently train in a precisely modeled and identified rigid body simulation with off-policy deep reinforcement learning, significantly sped up by a domain adapted curriculum, leading to a moderate 600 CPU hours of training time. The resulting policy is robustly transferred to the real humanoid DLR Hand-II, e.g., reaching more than 46 full $2 \pi$ rotations of the cube in a single run and allowing for disturbances like different cube sizes, hand orientation, or pulling a finger.
\end{abstract}

\section{Introduction}

\begin{figure}[t]
	\centering
	\includegraphics[width=0.6\linewidth]{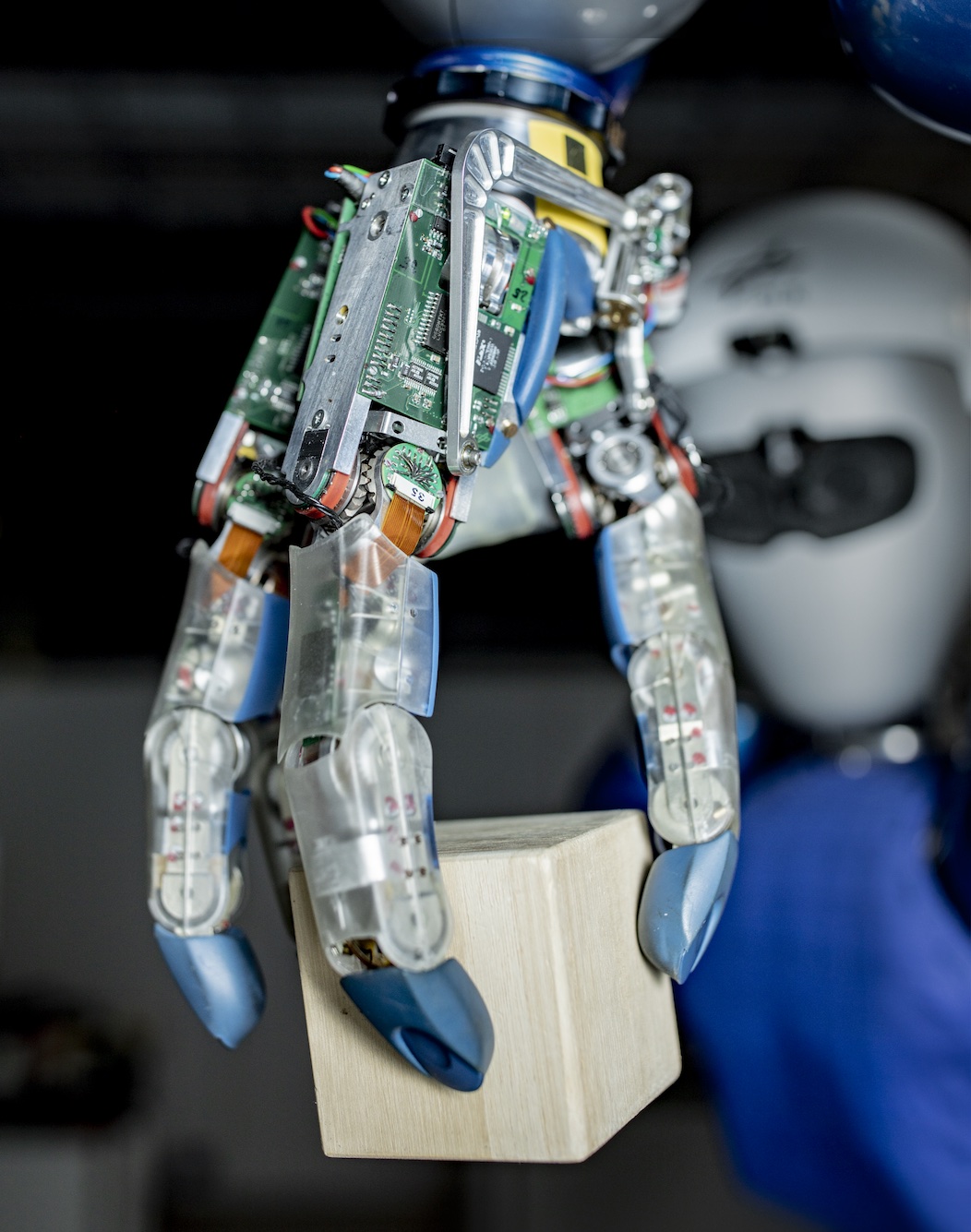}
	\caption{DLR's humanoid robot Agile Justin~\cite{Bauml2014} performing an in-hand manipulation task with the torque-controlled DLR Hand-II~\cite{Butterfass2001}: rotate the cube without dropping it. Only the built-in joint angle and torque sensors of the hand are used but no cameras.}
	\label{fig:justin_cube}
\end{figure}

Dextrous in-hand manipulation, i.e., moving and reorienting an object inside the hand without dropping it (see \cref{fig:justin_cube}), is a challenging task demanding for complex multi-finger strategies with intricate multi-contacts.
This is even more so when the task has to be performed robustly with permanent force closure, e.g., to withstand gravity in an upside-down setting.
Often the task has to be executed blindly, e.g., because of occlusions by the hand itself, only based on tactile feedback.
Humans achieve all of this with ease, using in-hand manipulation all the time in everyday live.

\subsection{Related Work}

For simulated environments, there is a large body of work investigating in-hand manipulation for challenging and dynamic tasks, some of which are using advanced humanoid hands.
The methods applied range from classical control and planning methods~\cite{Mordatch2012, Bai2014} up to learning from scratch applying modern deep reinforcement learning algorithms~\cite{Rajeswaran2017, Barth-Maron2018, Li2020}.
In these works, the full dynamic state of the manipulated object is taken for granted which is easy to achieve when working in simulation.

To use the full object state also on a real robotic system, an additional visual tracking system has to be added.
In their seminal work, OpenAI~\cite{OpenAI2018} used visual tracking not only for the object state but also for the finger tips. This way and in addition with domain randomization they could achieve robust sim2real transfer of a complex manipulation strategy, which was learned in simulation, to a real five-finger hand (Shadow Hand). The task was to rotate a cube  to given orientations in the hand with the palm facing up, hence, without the need for force closure. Using on-policy deep reinforcement learning, the task was learned from scratch but needed a huge compute budget of about 300k CPU hours (about 30 CPU years). Using the same robotic and learning setup, in a follow-up paper~\cite{OpenAI2019} they even learned to solve a (sensor-equipped) Rubik's cube single-handedly, but again without permanent force closure and needing an extremely high compute budget of 10k CPU years.

\citet{Haarnoja2018b} and \citet{Nagabandi2020} are also using visual object tracking, but directly learn on the real robotic system. The former solves the task of rotating a valve with their off-policy sample efficient Soft Actor Critic (SAC) algorithm and the latter uses dynamic learning and model predictive control to manipulate two Baoding balls. Both tasks are not force closure.

Other work uses no explicit object state but only tactile information (via tactile sensing or via torque controlled joints), e.g., in a model-based control setup~\cite{Fearing1986} or with (reinforcement) learning directly on the robotic system~\cite{Hoof2015, Kumar2016, Falco2018, Li2014}. All these in-hand manipulation tasks naturally use force closure but are all rather simple, i.e., small movements of the object without regrasping.

Recently, \citet{Bhatt2021} presented an approach for in-hand manipulation without any sensor feedback with their intrinsically compliant (pneumatic) hand. By executing carefully hand-crafted skills (e.g., rotate a cube by $\pi/2$) in sequence, they can perform similar tasks as the cube manipulation task from OpenAI. However, the execution of a long running task with regrasping which needs permanent force closure, like our upside-down cube rotation task, is not demonstrated. We show in \cref{sec:results} that open-loop replay of a hand-crafted policy gets unstable over time (e.g., the cube moves slowly up- or downwards), whereas learning can come up a stable closed-loop controller.

\subsection{Contributions}

We show for the first time that a policy for an in-hand manipulation task can be learned and robustly executed on an advanced humanoid hand solving the following challenges:
\begin{itemize}
	\item It uses no direct/external observation of the object state but is purely based on built-in position and torque sensing (tactile).
	\item It performs regrasping while demanding force closure at all times. This also makes learning harder as exploration motions are not allowed to break the force closure.
	\item The purely tactile task is learned from scratch in simulation and transferred to the real robot system (to our knowledge such a sim2real transfer for a tactile task has never been shown before).
	\item The training is efficient with a compute budget of CPU days instead of CPU years.
\end{itemize}

\section{Robotic System \& Simulator}

\subsection{The DLR Hand-II}
\begin{figure}[t]
	\centering
	\vspace{2mm}
	\includegraphics[width=0.52\linewidth]{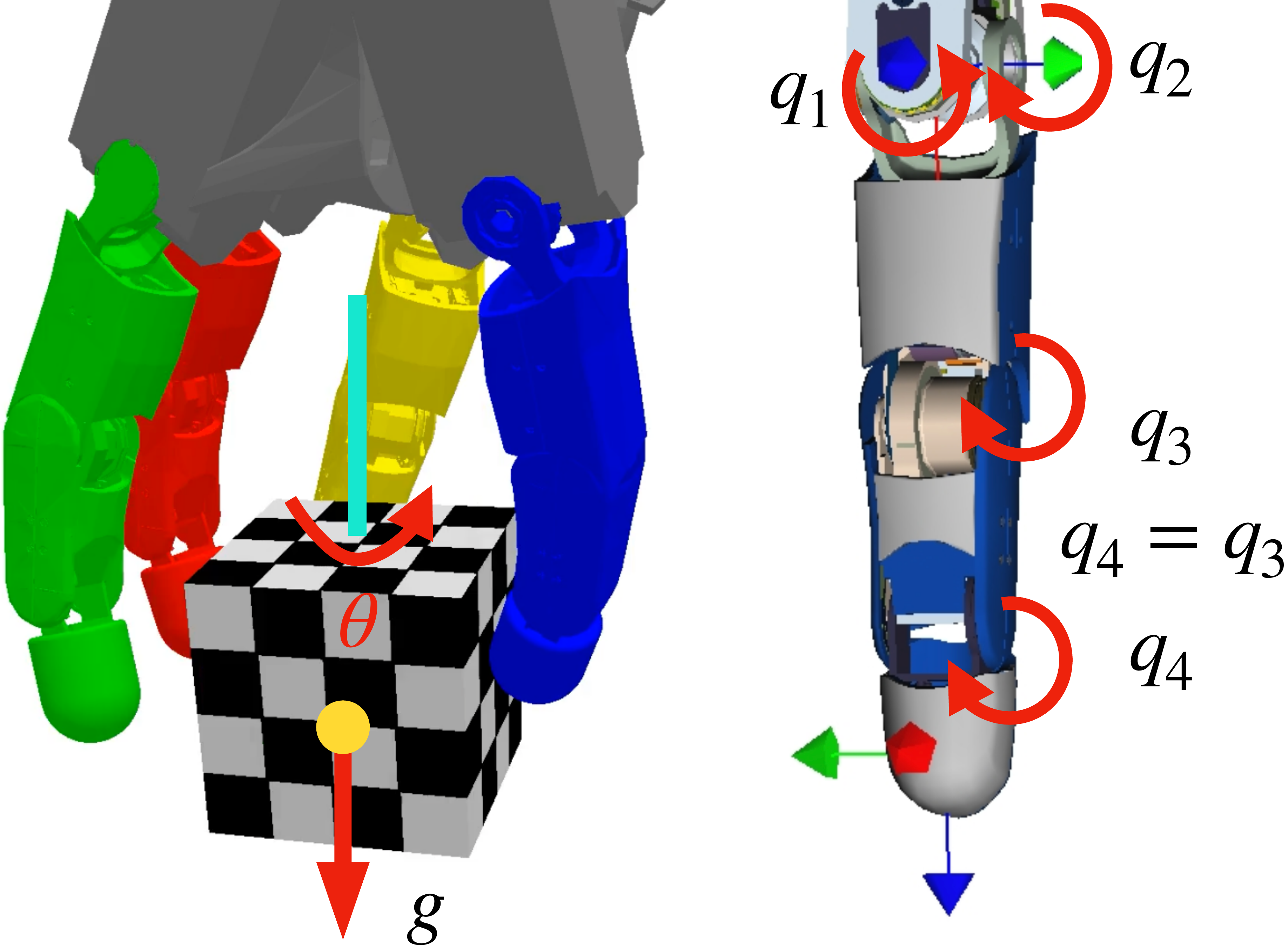}
	\vspace{-2mm}
	\caption{Left: The geometric model in the PyBullet rigid body simulator. The hand base (grey) connects the ring (yellow), middle (red), forefinger (green), and thumb (blue). Right: Each finger has three actuated and one passive DOF, with the constraint $q_4 = q_3$ enforced by a tendon drive. The angle $\theta$ indicates the cube rotation around the vertical axis.}
	\label{fig:hand_sim}
\end{figure}

The DLR Hand-II~\cite{Butterfass2001} features four fingers with three actuated DOFs each, as shown in Fig.~\ref{fig:hand_sim}.
Every joint is equipped with an output-side torque sensor which is used by a torque-controller, hence, controlling the output-side torque. This not only allows for accurate and stable grasping but also for a precise and meaningful simulation model of the hand as many complex effects in the drivetrain, esp. slip-stick friction, are dealt with by the torque-controller.

\subsection{Simulator}
We built a rigid body simulation in PyBullet~\cite{Coumans2016}. As it is open source it allows detailed insights and modifications of the simulation engine.
The required geometries for the hand were taken from CAD files. Only the fingertips are convex while the other links are not, but they are convexified in PyBullet leading to slight geometric errors.
We use lateral and spinning friction for the contact between the cube and the fingertip, but almost no friction between cube and all other links to force the manipulation to be mainly performed with the fingertips.
\subsection{Dynamics}
\begin{figure}[t]
	\centering
	\vspace{2mm}
	\includegraphics[width=0.78\linewidth]{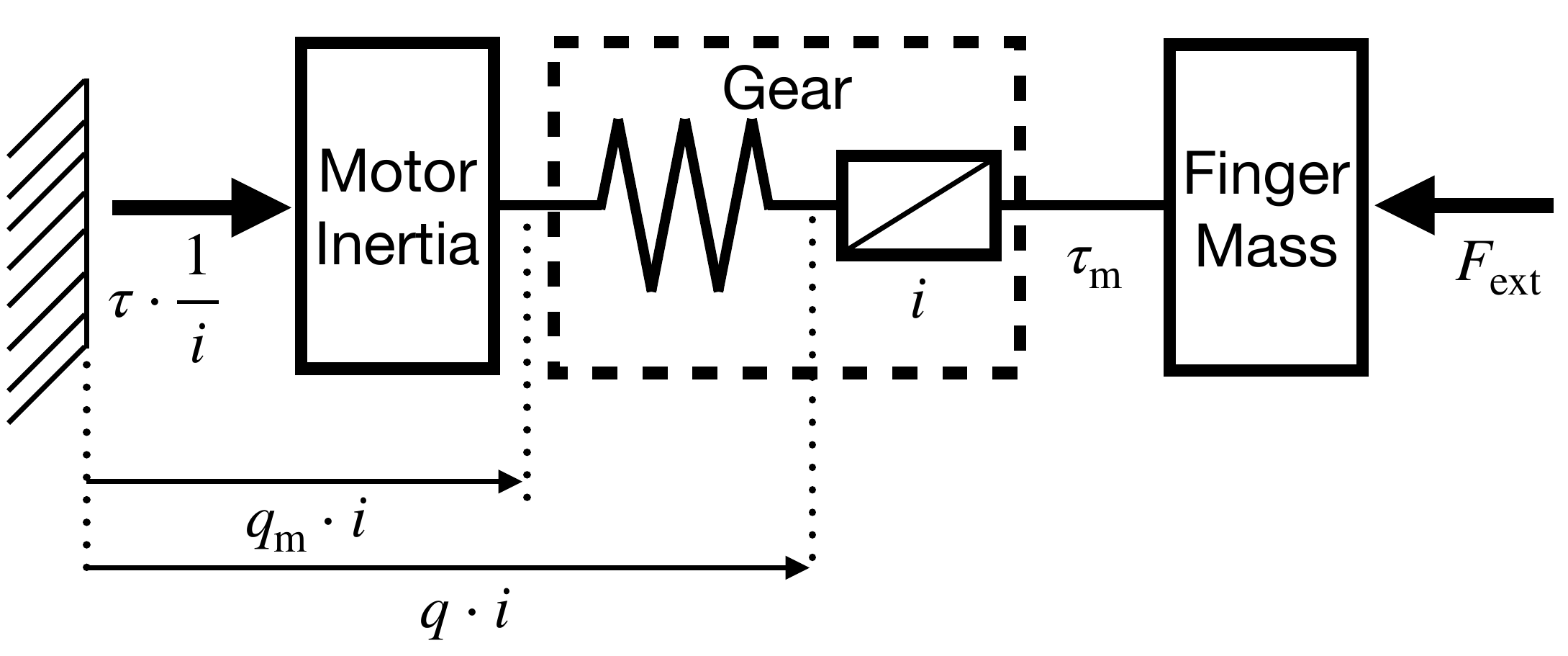}
	\vspace{-2mm}
	\caption{The drivetrain of the DLR Hand-II. The motor is equipped with an incremental position sensor. A strain gauge based sensor measures the output-side torque. The gear is modeled to be a motion ratio $i$ and a parasitic elasticity in series.}
	\label{fig:powertrain}
\end{figure}

The simulator is interfaced directly on the torque level.
We are modeling the drivetrain as shown in Fig.~\ref{fig:powertrain}.
To increase numerical stability, the whole assembly is transformed to the output (right) side of the gear, dividing the rotation speed of the inertia by the gear ratio $i$.
The applied torque $\tau$ is calculated as
\begin{equation}
	\tau = K_\text{P} (q_\text{d} - q_\text{m}) - K_\text{D} \dot{q},
\end{equation}
with $q_\text{d}$ denoting the desired position. The measured position is calculated as
\begin{equation}
	q_\text{m} = q - \frac{\tau}{K_\text{e}},
\end{equation}
with the true joint angle $q$ and the parasitic stiffness $K_\text{e}$.
The joint angle velocity $\dot{q}$ is taken directly from the simulator state, whereas on the real system  a low-pass filter is required to calculate the position's time derivative because of sensor noise.
$K_\text{P}$ and $K_\text{D}$ indicate the PD controller parameters.
The whole simulation loop runs at $f_\text{sys}=500$\,Hz.

\subsection{System Identification}
\label{subsection:system_identifiction}
One source of discrepancies between simulation and real world are the geometric errors due to inaccuracies in the production and backlash. We abstract these errors into joint angle measurement offsets. To identify an appropriate range for sampling offsets we mirror the measured joint angles of the real system in the simulation and move fingers towards each other until they touch. The resulting discrepancy is taken as the maximal joint angle offset.

We also identify the parameters governing the dynamics and the contact behavior of the simulation.
Note that we are assuming calibrated torque and position sensors.
All sensors are deterministically initialized during startup of the robot.
\begin{description}
	\item[$K_\text{P}$] The proportional gain of the position controller can be verified by measuring the force exerted at a given link position. The finger has to be static while the control error is constant.
	\item[$K_\text{e}$] As seen in Fig.~\ref{fig:powertrain}, the parasitic stiffness of the drivetrain can be estimated by fixing the articulated links relative to the hand's base. When the motor applies a torque $\tau$, the motion $\Delta q_\text{m}$ can be measured. The stiffness is then approximated using the relation $K_\text{e} = \tau_\text{m} / (\Delta q_\text{m} + \epsilon)$, with a small $\epsilon$.
	\item[$K_\text{D}$] The damping coefficient of the PD controller subsumes different effects like joint friction and controller parameters  into a single variable. It can be determined by matching an open-loop trajectory to pre-recorded data from the real hand.
	\item[$F_\text{s}$] The static friction of the links can also be determined by matching open-loop trajectories.
\end{description}

\begin{figure}
	\centering
	\vspace{2mm}
	\includegraphics[width=0.78\linewidth]{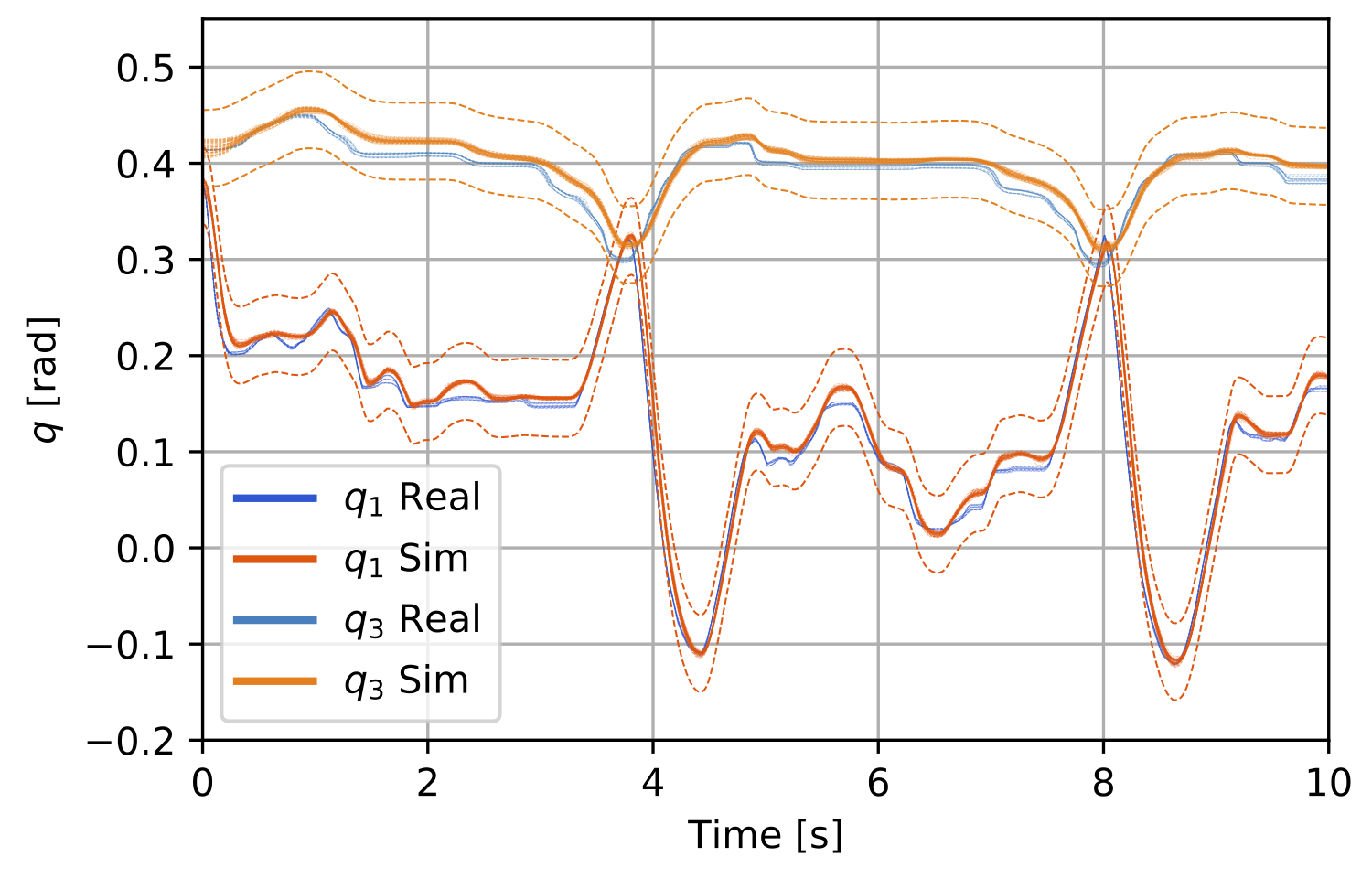}
	\vspace{-3mm}
	\caption{
		Open loop trajectories of two ring finger joint angles $q_1$ and $q_3$ following previously recorded target positions of a trained policy.
		In blue we show $15$ individual real trajectories across three robot initializations, and in red $50$ individual simulation trajectories each with dynamics sampled from the domain randomization distribution used for training.
		The dashed lines indicate the range of the geometric randomization.}
	\label{fig:dynamics}
\end{figure}

Fig.~\ref{fig:dynamics} shows the distributions over the real dynamics and the simulated one for typical joint angle trajectories of a freely moving finger (no contacts). One can see that the real dynamic is highly deterministic and it would be possible to fit the simulated dynamic even better with a more complex model. However, the plot also shows that compared to the randomization required to cover the geometric uncertainty the dynamics play a minor role. That is, of course, largely due to the fact that we intentionally move the fingers slowly to make the sim2real transfer more robust.

For the rigid body simulation, we calibrated the lateral as well as the spinning friction between fingertips and the cube by recording the required torque to hold a defined load.
Experiments show that even this crude approximation suffices to allow a sim2real transfer.

\subsection{Control Architecture}
\begin{figure}[t]
	\centering
	\vspace{3mm}
	\includegraphics[width=.88\linewidth]{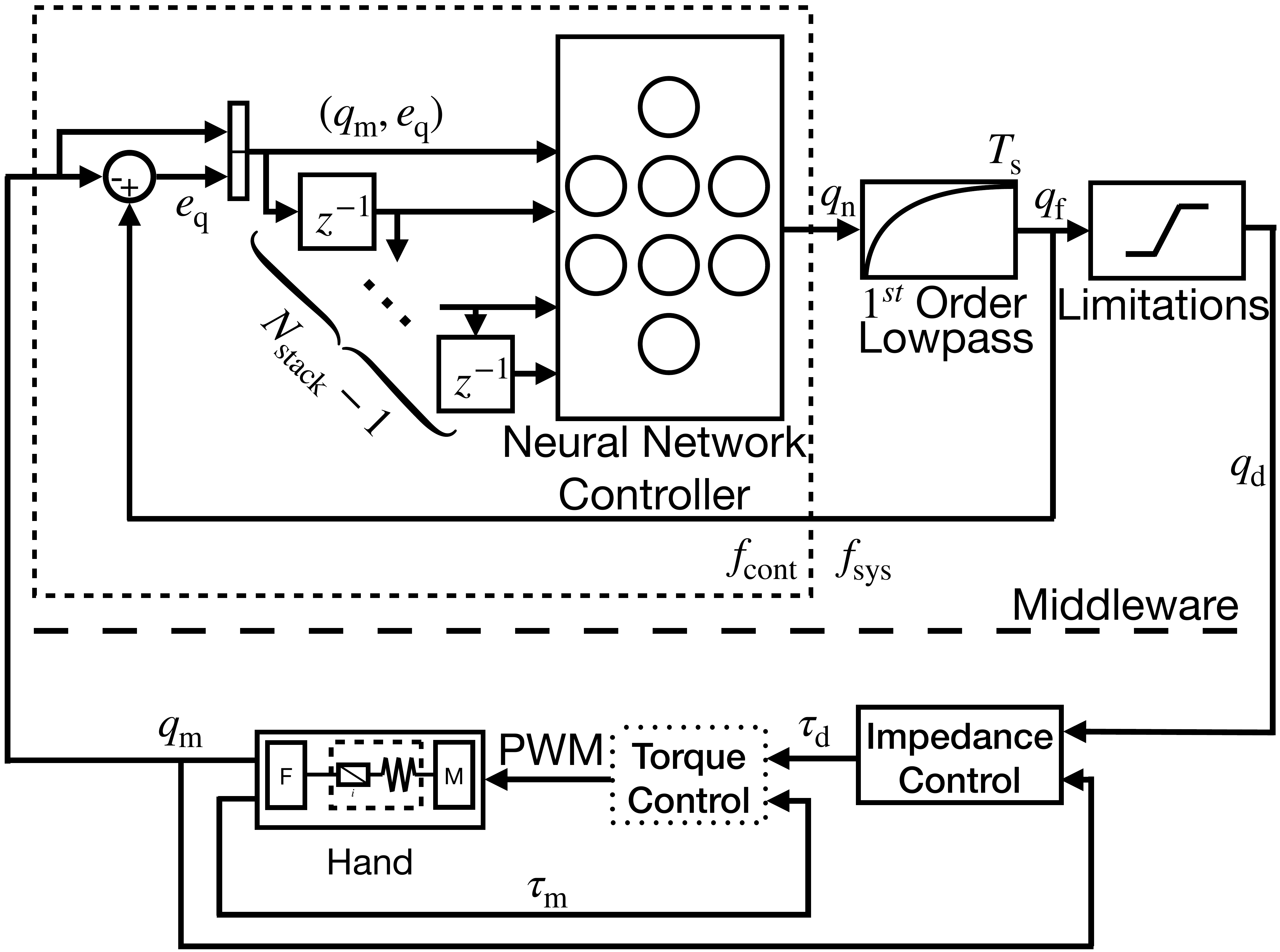}
	\caption{Control architecture including the learned policy. The network receives the last $N_\text{stack}$ joint angles as well as control errors. It outputs the desired angles $q_\mathrm{n}$ for each joint of the hand, which then get filtered and clipped (to avoid hand damage). The observation collection and the network run at a rate of $f_\mathrm{cont} =10\,\mathrm{Hz}$), whereas the other parts of the control loop run at $f_\text{sys}$ (1\,kHz on the robot and 500\,Hz in simulation). The torque controller is only present on the real system since the desired torque can be set directly in simulation. The impedance controller is implemented as a PD controller.}
	\label{fig:network}
\end{figure}

As shown in Fig.~\ref{fig:network}, the neural network controller calculates a new target joint angle $q_\text{n}$ at a rate of  $f_\text{cont} = 10$\,Hz.
A first order low-pass filter with a cutoff frequency of $1/T_\text{s} = 2$\,Hz smoothens the signal at a rate of $f_\text{sys} = 1000$\,Hz on the real hand and $f_\text{sys} = 500$\,Hz in the simulation, giving the signal $q_\text{f}$ as output.
After limiting the rate of change as well as the resulting torque (by approximate inversion of the controller), the value is fed to the lower level hand controller as $q_\text{d}$.
As input, the neural network controller receives the measured joint angle $q_\text{m}$ and the control error $e_\text{q} = q_\text{f} - q_\text{m}$ of the current as well as the last $N_\text{stack} - 1$ executions.
To increase the robustness of the controller, we do not provide any direct velocity (inferable from the position history) or torque (encoded in control error and velocity) feedback.

\section{Learning the Manipulation Task \protect\footnote{See the project website for the learning and simulation parameters.}}

\subsection{Learning Algorithm}
To learn the weights of the neural network controller we use reinforcement learning (RL). In particular, we use the Soft Actor-Critic (SAC) algorithm introduced by~\citet{Haarnoja2018}.
We chose SAC because it is a simple algorithm with impressive performance on standard benchmarks with much fewer environment steps compared to on-policy algorithms such as Proximal Policy Optimization (PPO) \cite{Schulman2017} as, e.g., used in OpenAI's manipulation work~\cite{OpenAI2019}.

\subsection{Task Specification}
The goal of the RL agent is to hold a cube from above and rotate it as far as possible around the $z$-axis without dropping it.
Episodes start with the fingers located closely around the cube without fully grasping it.
The cube does not experience gravity for a short time initially ($< 0.5$\,s), which corresponds to a human passing the object into the robots hand.
During training, episodes are reset when the cube drops, moves away too far in the plane, or a certain time limit is reached (usually between 10\,s and 20\,s).

\subsubsection{Reward}
The agent is rewarded proportional to the rotation achieved in the last time step and punished for moving or rotating the cube along any other axis.
\begin{equation*}
	R(t) = \lambda_\mathrm{spin} \Delta_\theta - \lambda_\mathrm{z} |\Delta_z| - \lambda_\mathrm{plane} |\Delta_{xy}| - \lambda_\mathrm{rot} (|\Delta_\psi| + |\Delta_\phi|)
\end{equation*}

Additionally, we punish the agent in the last step if the episode terminates due to the cube going out of the allowed translational range.

\subsubsection{Observations}
While the controller network, i.e. the policy $\pi$, receives only the stacked joint angles and control errors as explained above, the $Q$ network is provided with additional information only available in simulation. This is common practice \cite{OpenAI2018} \cite{OpenAI2019} and can facilitate faster learning and more accurate $Q$ value estimates. We pass:
\begin{itemize}
	\item cube position [$x, y, z$]
	\item cube orientation [$\sin(\theta), \cos(\theta), \phi, \psi$]
	\item cube linear velocities [$\dot{x}, \dot{y}, \dot{z}$]
	\item cube rotary velocities [$\dot{\theta}, \dot{\phi}, \dot{\psi}$]
	\item the last control target [$q_\mathrm{f}$]
\end{itemize}
Because the angle~$\theta$ increases arbitrarily it is disambiguated by passing $\sin \theta$ and $\cos \theta$, whereas the other angles $\phi$ and $\psi$ stay near 0. To speed up learning, we exploit the symmetry of the cube by using $4 \cdot \theta$ instead of $\theta$ in all our experiments.
Also note that the last control target is redundant information and we use observations stacking with $N_\text{stack} = 5$.

\subsubsection{Challenge}
\begin{figure}
	\centering
	\vspace{2mm}
	\includegraphics[width=0.93\linewidth]{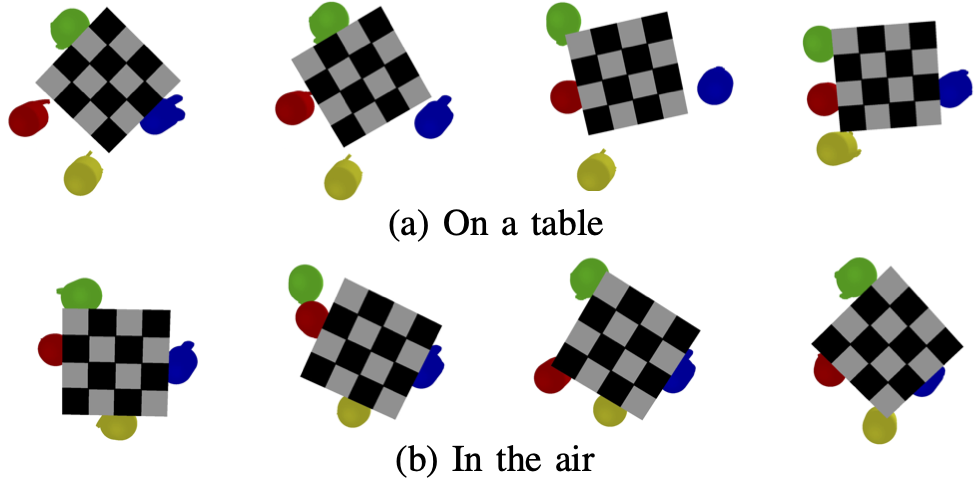}
	\vspace{-2mm}
	\caption{Sequences (images $400$\,ms apart) of the hand rotating a cube in different settings. Only the last link is rendered. Scene is shown from below. The required force closure in (b) restricts the agent to more careful actions.}
	\label{fig:sequence}
\end{figure}

In contrast to the manipulation of objects lying on a table or the palm of the hand, the main difficulty of our task for current state-of-the-art RL algorithms is to explore new actions while maintaining force closure (compare Fig.~\ref{fig:sequence}).
Only a single small mistake is enough to lose control and drop the cube, ending the episode.
However, limiting the exploration noise makes learning regrasping hard because to choose a finger to lift off and move to another side is difficult as it requires highly correlated actions over a reasonably long time frame.
That leads to certain local optima, which are hard to escape.

\begin{figure}
	\centering
	\vspace{2mm}
	\includegraphics[width=.86\linewidth]{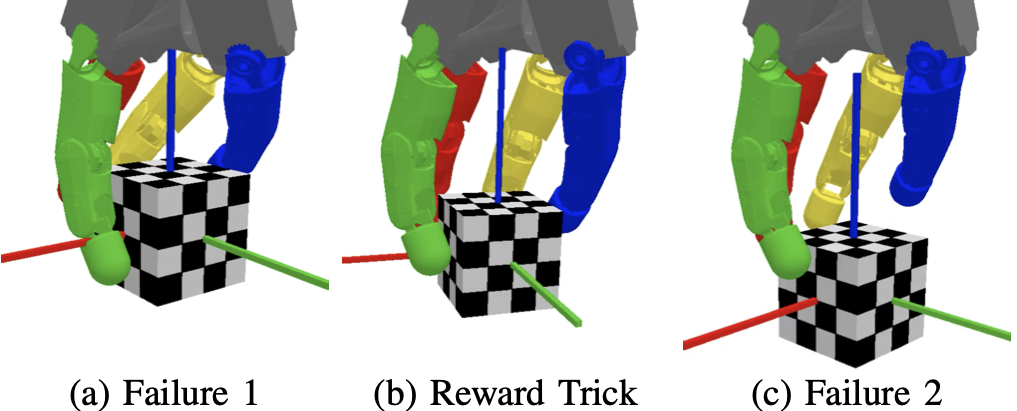}
	\vspace{-2mm}
	\caption{The most common failure cases (a) and (c) compared to the preferred behavior (b). In both failure cases the agent rotates the cube approximately $\pi / 2$. (The green axis initially points towards the green finger.) In (a) the agent gets stuck with badly twisted fingers. Adding a penalty term for $q_1$ prevents this, leading to more natural finger positions (b). In (c) the agent opens the fingers at the end of the twisting motion, so that the cube rotates a little bit further but drops immediately.}
	\label{fig:failure}
\end{figure}

In particular, the failures in ~\cref{fig:failure} are not posing a problem if the cube is resting on a table, because then the agent can freely move its fingers away from the cube after rotating it as far as the kinematics initially allows.

\subsection{Increasing Training Efficiency \& Sim2Real Robustness}
\label{subsection:reward}
Specifically to prevent the first failure case in Fig.~\ref{fig:failure} we added an additional penalty term to the task reward.
This penalty term increases the success probability and allows for a larger range of initial conditions (e.g. the initial $z$ position of the cube relative to the hand) that can be successfully trained for.
Therefore, all experiments reported in this paper have been conducted with this penalty term.
Concretely, we add the following reward term:
\vspace{-2mm}
\begin{equation*}
	R_\text{proxy}(t) = -\lambda_\text{proxy} \sum_n^{N_\text{fingers}} ({q_1^{(n)}})^4
\end{equation*}
at each time step. Where $q_1$  is the joint that moves the finger left or right (cf. Fig.~\ref{fig:hand_sim}).
The effect of this penalty can be observed in Fig.~\ref{fig:failure}. Not twisting the finger too far allows the agent to learn by sliding the fingers around.

We also clip the maximum available rotation reward per step such that rotations faster than some constant (usually 1\,rad/s) are not rewarded.
That is a very moderate speed, which is well suited for sim2real transfer and forces the agent to focus on stability across the randomized domain instead of spinning ever faster.

\subsection{Curriculum}
We used curricula and training continuation for policies in slightly more difficult environments multiple times, before finding the parameters that allow for efficient learning in simulation and robust sim2real transfer.

Using these parameters, our training runs that learn from scratch still include a curriculum to increase the filter constant $T_\mathrm{s}$ and gravity, because we found that large $T_\mathrm{s}$ in combination with full gravity hinders exploration.
Interestingly, shortly after our initial submission \citet{Chen2021Isaac} independently showed that a gravity curriculum was necessary to learn to manipulate different objects in simulation with an upside-down Shadow Hand.

Table~\ref{table:curriculum} shows approximately how many million environment steps each run required to solve the task, i.e. started regrasping. We ran three different settings each with high and low filter constant for 6.5 million steps.
"Table" refers to the task where the cube sits on a table and shows how much easier the problem becomes if continuos grasping is not a requirement.
"Float" refers to the task where the cube's acceleration due to gravity is reduced by $95\%$.
"Full" refers to the task with full gravity and no helping table.

While this is only a small sample size, there is a tendency that learning with full gravity and high filter constant is the least successful setting, and either speeding up the dynamics of the hand or slowing down the dropping of the cube improves the results. Moreover, not visible in the table is that the two runs with full gravity and high filter constant that did not succeed after 6.5 million steps ended up in the local optimum where the agent spins and drops the cube immediately (which will likely never recover, cf. Fig.~\ref{fig:failure}), while all other failed runs are still holding and wiggling the cube.

\begin{table}
	\centering
	\vspace{3mm}
	\begin{tabular}{c  c c c c c c}
		\hline
		             & \multicolumn{3}{c}{$T_\mathrm{s} = 0.5$\,s} & \multicolumn{3}{c}{$T_\mathrm{s} = 0.1$\,s}                               \\
		\hline
		Run          & Table                                       & Float                                       & Full & Table & Float & Full \\
		\hline
		1            & 0.3                                         & 2.2                                         & 1.8  & 0.4   & 1.5   & 5.1  \\
		2            & 0.4                                         & 1.7                                         & -    & 0.2   & 5.8   & 1.7  \\
		3            & 0.3                                         & 0.9                                         & -    & 0.2   & -     & 6.0  \\
		4            & 0.4                                         & -                                           & 2.3  & 0.2   & 2.4   & 3.9  \\
		\hline
		Success [\%] & 100                                         & 75                                          & 50   & 100   & 75    & 100  \\
		\hline
	\end{tabular}
	\caption{Time constant \& gravity}
	\label{table:curriculum}
\end{table}

\begin{figure}
	\centering
	\includegraphics[width=0.9\linewidth]{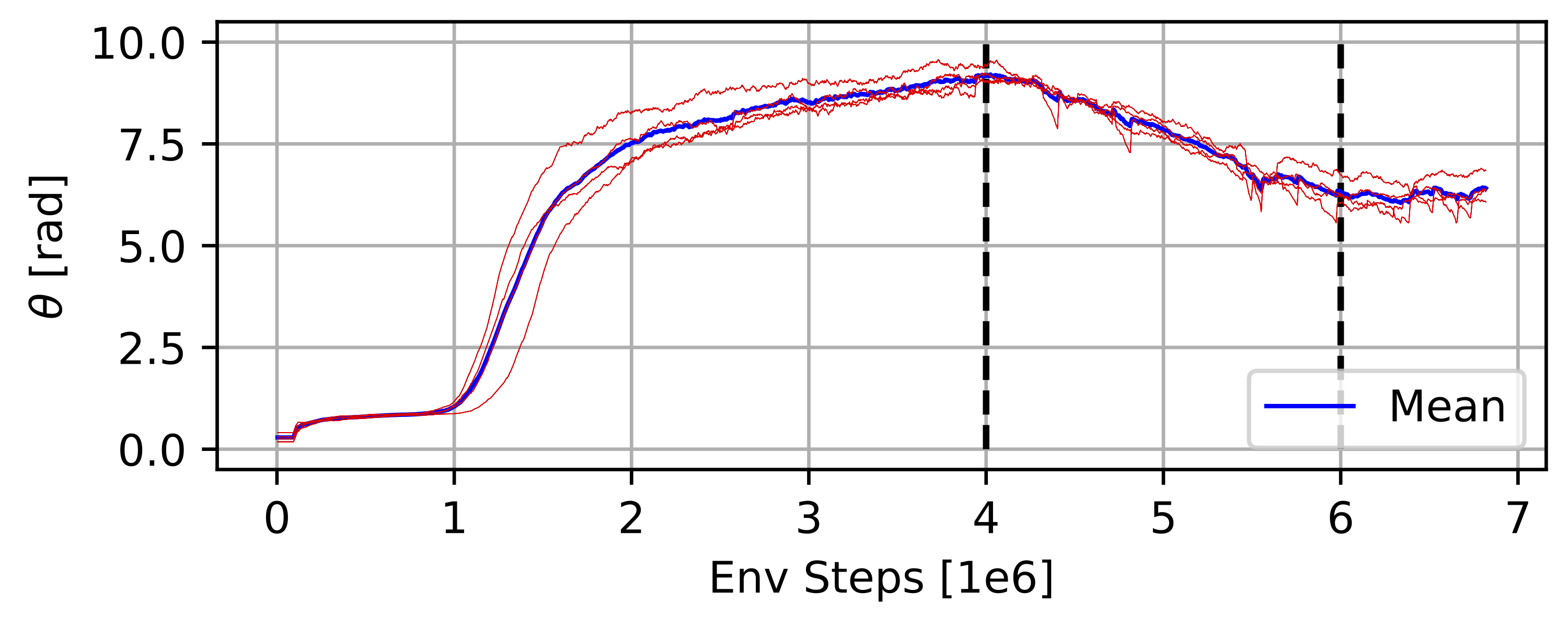}
	\vspace{-3mm}
	\caption{
		Typical learning curves with curriculum.
		We show the average of the final angle $\theta$ over the training episodes (this metric correlates strongly with the reward and is easy to interpret) for four individual runs.
		Values are filtered with exponential moving average over 50 steps.
		The dashed black lines indicate the beginning and end of the curriculum.
		Initially, all agents quickly learn to rotate approximately $\pi / 2$.
		After around one million steps the agents learn to "regrasp" and quickly gain performance (Note: often it takes much longer for some seeds than others).
		Between two and four million steps they slowly improve their performance.
		Then the curriculum sets in and slows down the agent.
		After the curriculum finishes at six million steps the agents start to slowly improve again.
	}
	\label{figure:learning_curve}
\end{figure}

\cref{figure:learning_curve} shows a batch of four runs trained with identical settings as the experiments in  \cref{table:curriculum} except the added curriculum, which linearly increases the filter constant from $T_\mathrm{s} = 0.12$\,s to $0.5$\,s and the gravity from effectively $5\%$ to $100\%$.
These agents are almost ready to be run on the real system.
To reach the full sim2real potential we fine-tune them one more time in a slightly more difficult simulation with longer episodes, lower entropy target and higher discount factor ($\gamma_\text{inital} < \gamma_\text{fine})$, to prime them for longer evaluations).

\subsection{Randomization}
To enable proper sim2real performance we randomize various aspects of the simulation.
We randomize the constants identified in Section~\ref{subsection:system_identifiction}.
As mentioned before, we found that the randomization of the geometry is more important than the dynamics for our task.
Moreover, we add gaussian noise to the sensor readings (joint angle measurements), and $10\%$ sticky actions as in the Atari benchmarks~\cite{Machado2018} to compensate for potential timing errors.
Finally, we randomize the size, mass and initial pose of the cube, as well as the initial joint angles of the hand.

\subsection{Implementation}

\subsubsection{Networks}
We use a fully connected network with two layers, each with 512 neurons for all of our experiments.
This simple architecture allows for training  on the CPU.

\subsubsection{Parallelization}
Although SAC was designed to learn on the real system and, therefore, a single environment, it is possible to collect data with multiple workers.
We use 12 workers for all our experiments and found this parallelization to provide a significant speed up.
Additionally, we run the learner on data sampled from the replay buffer's state of the previous iteration allowing us to fully parallelize learning and data collection, while maintaining fully deterministic results.

\subsection{Discussion}
The sample efficient algorithm and the parallelization allow us to train a fully functional controller from scratch in less than 36 hours on a 16 Core desktop machine, hence only moderately 600 CPU hours are required.
That is surprising compared to the 300k CPU hours required by \citet{OpenAI2018} to learn manipulation of a cube in the palm of the Shadow Hand.
However, \citet{Plappert2018DDPG} show that a very similar tasks, can be learned using the DDPG+HER algorithm with a compute budget similar to ours (based on a follow-up paper \cite{Melnik2021}, we estimate about 200 CPU hours), although they do not prove sim2real transferability.
Since DDPG and SAC are both sample-efficient "value gradient" algorithms, we attribute the largest gains in our training time to the choice of the algorithm.

Quantifying how much each of our modifications (4x symmetry, joint angle penalty, and curriculum) contribute to the learning success is not easy. Measuring training times for a few runs is no problem but verfying that some tweak improves the range of "learnable" initial conditions would take hundreds of runs.

As mentioned in~\ref{subsection:reward}, from all our experiments over several months, we are confident that the joint angle penalty does make the learning more robust to environment modifcations, and therefore sped up our development process.

The effect of the curriculum was easier to measure and is quantified in Table~\ref{table:curriculum}. It shows that curricula can be a useful tool to bring current RL algorithms closer to real world applications.

Finally, we recommend that symmetries should be exploited for real world applications whenever possible. Not exploiting them in our experiments leads to steps in the learning curves, because the agent enters completely unexplored states in the observation space.
Usually, the agent solves the following three subproblem faster than the first one, presumably because the networks learn to generalize between the symmetries.

\section{Results on the real Robot}
\label{sec:results}

\begin{figure}
	\centering
	\vspace{2mm}
	\includegraphics[width=0.95\linewidth]{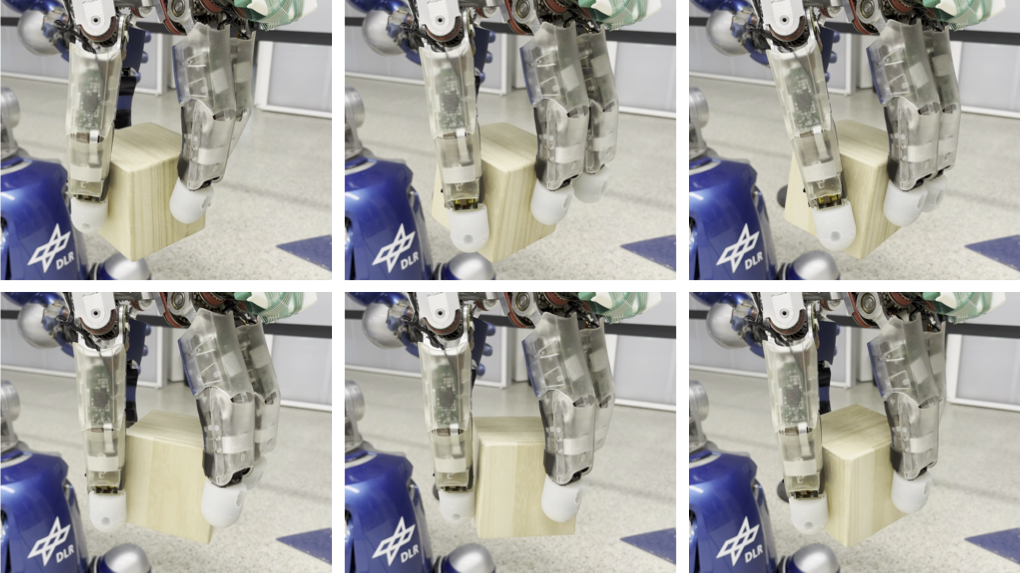}
	\vspace{-1mm}
	\caption{The real hand turning the cube (frames 880\,ms apart).}
	\label{fig:real_sequence}
\end{figure}

\cref{fig:real_sequence} shows a successful in-hand manipulation experiment using a policy trained in simulation. For all reported experiments in this section, we use a policy that was trained for approximately $36$ hours as described in the previous section. The accompanying video shows multiple runs of this policy as well as other policies, demonstrating the robustness of our approach.

\subsection{Experiments}
\subsubsection{Quantitative Result}
To quantify the robustness of the policy and the sim2real transfer, 10 consecutive runs were performed with a human passing the cube freehand (no mechanical guidance) to the robot leading to variations in the initial cube pose.
A run was stopped after $40$\,s (double the time horizon that the policy was trained on) or before, if the cube dropped or got stuck. A run is regarded successful if it reached at least $80$\,\% of the best performing run ($\theta = 15\,\mathrm{rad}$). In result, 8 of the 10 runs on the real robot were successful. In comparison, in the simulation 9 of 10 runs were successful, although the variation in the initial cube pose was smaller.

\subsubsection{Stability}
We tested how long and far the policy could rotate the cube with the real hand. After nearly $16$ minutes and $46$ full rotations ($\theta = 289\,\mathrm{rad}$) we eventually stopped the experiment to limit the tear on the robot hardware.

\subsubsection{Open-Loop Control}
To show that the policy actually closes the control loop using the sensor feedback instead of just executing a simple, almost constant pattern, we performed an additional experiment. In a successful try, the trajectory of the desired joint angles was recorded and then replayed for a cube in a very similar initial pose. This trajectory replay already failed after $\theta=\pi/2$, whereas using the active policy with feedback control leads to runs with more than $\theta=289\,\mathrm{rad}$.

\subsubsection{Hand-Crafted Trajectory}
For this experiment, we hand-crafted a trajectory for the fingers to rotate the cube, similar to what was done by \citet{Bhatt2021} for their flexible pneumatic hand. This trajectory worked but reached only a $\theta = 11\,\mathrm{rad}$ before dropping the cube compared to the more than $\theta = 289\,\mathrm{rad}$ using the learned closed-loop  policy.

\subsubsection{Perturbations}
\begin{figure}
	\centering
	\vspace{2mm}
	\includegraphics[width=0.95\linewidth]{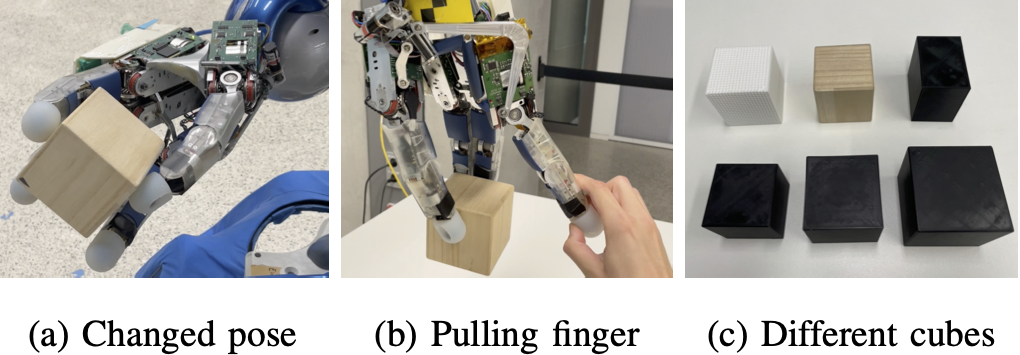}
	\vspace{-1mm}
	\caption{Left and middle: Successful manipulation while exposed to disturbances unseen during training. Right: Different cube sizes (variation of 2\,cm), materials (wood, plastic, nubby plastic) and aspect ratios the policy can handle.}

	\label{fig:perturbation1}
\end{figure}

The controller is able to deal with previously unseen perturbations on the real system like pulling a finger or changing the hand's orientation as shown in Fig.~\ref{fig:perturbation1}.
Policies are also able to handle cubes with different sizes of up to 2\,cm variations. When fine-tuning a policy by training continuation, even the manipulation of cuboids can be achieved.

\subsection{Discussion}

The high success rate as well as the large achievable rotation angle proves the high performance of the learned policies and the robustness of the sim2real transfer.

Taking a closer look at the execution of the policy, a typical behavior is to "try again" to put a finger on an edge after a previous attempt failed while regrasping. Our interpretation of this behavior is that using the control error of the PD controller, which is provided as an input, the policy can "feel" the failure state and react accordingly. This is a direct consequence of the availability of the link-side torque-sensors as only based on them, the high-fidelity impedance controller could be implemented, which the policy outputs the desired joint angles to.

In the hand-crafting experiment we also take advantage of the impedance control (hence, the torque sensors) to control the finger motion in free space as well as the forces in contact via the desired joint angles in a unified way. However, the bad performance in this experiment shows that it is almost impossible to manually come up with joint angle trajectories that avoid the slow moving up or down of the cube, which eventually leads to dropping it. But the policy is learning such a behavior during training.

\section{Conclusions}
We have shown that the challenging in-hand manipulation task of rotating a cube upside-down, without any external (e.g., visual) tracking of the object state, can be efficiently learned from scratch and robustly executed on the advanced torque-controlled DLR Hand-II (more than 46 rotations and resilient to disturbances). For this, we built and identified a model of the hand in a rigid body simulator including torque-based impedance control and drivetrain elasticities. In addition, we significantly sped up SAC-based reinforcement learning by identifying a learning curriculum (filter~constant/gravity) and using domain knowledge.

This is the first time that such a complex purely tactile in-hand manipulation task with continued regrasping, requiring permanent force closure, was efficiently learned and reliably executed. In the future, we want to extend this method to more flexible manipulations, e.g., rotating around the x- and y-axes, and arbitrary but known geometries. This would open up in-hand manipulation as a standard tool in applications like service robotics or industrial manufacturing.

\bibliographystyle{IEEEtranN-modified}
\bibliography{IEEEabrv, research.bib}

\end{document}